\documentclass[runningheads]{llncs}
\usepackage{times}
\usepackage{graphicx}
\usepackage{latexsym}
\usepackage{times}  
\usepackage{helvet} 
\usepackage{courier}  
\usepackage[hyphens]{url}  
\usepackage{graphicx} 
\usepackage{graphicx}  
\usepackage{hyperref}
\usepackage{amsmath,amssymb,amsfonts}
\usepackage{algorithmic}
\usepackage{graphicx}
\usepackage{subfig}
\usepackage{textcomp}
\usepackage{xcolor}
\usepackage{multirow}
\usepackage[normalem]{ulem}
\useunder{\uline}{\ul}{}
\def\BibTeX{{\rm B\kern-.05em{\sc i\kern-.025em b}\kern-.08em
    T\kern-.1667em\lower.7ex\hbox{E}\kern-.125emX}}
\DeclareMathOperator{\E}{\mathbb{E}}
\begin{document}

\title{Learning more expressive joint distributions in multimodal variational methods
}
%
%

\author{Sasho Nedelkoski\inst{1}\and
Mihail Bogojeski\inst{2}\and
Odej Kao\inst{1}}
\authorrunning{S. Nedelkoski et al.}
%
\institute{Distributed Systems, TU Berlin, Berlin, Germany
\email{nedelkoski, odej.kao@tu-berlin.de}\\
\and
Machine Learning, TU Berlin, Berlin, Germany\\
\email{mihail.bogojeski@campus.tu-berlin.de}\\
}
\maketitle

\begin{abstract}
\sloppy{
Data often are formed of multiple modalities, which jointly describe the observed phenomena. 
Modeling the joint distribution of multimodal data requires larger expressive power to capture high-level concepts and provide better data representations. 
However, multimodal generative models based on variational inference are limited due to the lack of flexibility of the approximate posterior, which is obtained by searching within a known parametric family of distributions. 
We introduce a method that improves the representational capacity of multimodal variational methods using normalizing flows. 
It approximates the joint posterior with a simple parametric distribution and subsequently transforms into a more complex one. 
Through several experiments, we demonstrate that the model improves on state-of-the-art multimodal methods based on variational inference on various computer vision tasks such as colorization, edge and mask detection, and weakly supervised learning. We also show that learning more powerful approximate joint distributions improves the quality of the generated samples. The code of our model is publicly available at \href{https://github.com/SashoNedelkoski/BPFDMVM}{https://github.com/SashoNedelkoski/BPFDMVM}.}
\end{abstract}

\section{Introduction}\label{introduction}
\sloppy{ 
Information frequently originates from multiple data sources that produce distinct modalities. For example, human perception is mostly formed by signals that go through visual, auditory, and motor paths, and video is accompanied by text captions and audio signals. These modalities in separate describe individual properties of the observation but are also correlated with each other.} 

The learning of a joint density model over the space of multimodal inputs is likely to yield a better generalization in various applications. Deep multimodal learning for the fusion of speech or text with visual modalities provided a significant error reduction~\cite{potamianos2017audio,srivastava2012multimodal}. The best results for visual question answering are achieved by using a joint representation from pairs of text and image data~\cite{kim2016multimodal}. Distinct from the fully supervised learning where the mapping between modalities is learned, multimodal learning with generative models aims to capture the approximation of the joint distribution. This enables data generation from the joint as well as conditional distributions to produce missing modalities.

The complex nature of the multimodal data requires learning highly expressive joint representations in generative models. One of the most utilized approaches for multimodal generative models is variational inference. However, these models have limitations as the search space for the best posterior approximation is always within a parametric family of distributions, such as the Gaussian~\cite{wu2018multimodal,vedantam2017generative,suzuki2016joint}. Even when the complexity of the data increases, as multiple sources of information exist, the search space of the approximate posterior remains unchanged from the unimodal variant~\cite{kingma2013auto}. This type of parametric posterior imposes an inherent restriction since the true posterior can be recovered only if is contained in that search space, which is not observed for most data~\cite{berg2018sylvester}.

\emph{Contributions.} We propose an approach to learn more expressive joint distributions in the family of multimodal generative models, which utilizes normalizing flows to transform and expand the parametric distribution and learn a better approximation to the true posterior. 

First, we utilize the product-of-experts to prevent an exponential number of inference networks~\cite{hinton2002training}, leading to a reduced number of inference networks as one per modality. The output of the product-of-experts inference network is the parameters of the initial joint posterior distribution of the modalities, which is often Gaussian. Subsequently, we add a module that transforms this parametric distribution using continuous normalizing flows. This process produces a new and more complex joint distribution, enabling a better approximation to the true posterior in multimodal variational methods. The learned transformation allows the model to better approximate the true joint posterior. Samples from this distribution are subsequently used by the decoder networks to generate samples for each modality. 

We evaluate the quality of the method to the state of the art and show improvements on multiple different datasets. We demonstrate that the quality of the generated missing modalities is higher than that of the state of the art and that our model can learn complex image transformations. Furthermore, the experiments also show that the model can be used for weakly supervised learning with a small amount of labeled data to reach decent values of the loss function.



\section{Related Work}\label{relatedwork}

Recently, deep neural networks have been extensively investigated in multimodal learning~\cite{ngiam2011multimodal}. A group of methods, such as Deep Boltzmann Machines~\cite{srivastava2012multimodal} and topic models~\cite{barnard2003matching,jia2011learning,nedelkoski2019anomaly}, learn the probability density over the space of multimodal inputs (i.e., sentences and images). Although the modalities, architectures, and optimization techniques might differ, the concept of fusing information in a joint hidden layer of a neural network is common. 

Kingma et al.~\cite{kingma2013auto} introduced a stochastic variational inference learning algorithm, which scales to large datasets. They showed that a reparameterization of the variational lower bound yields a lower-bound estimator, which can be directly optimized using standard stochastic gradient methods. This paved the way for a new class of generative models based on variational autoencoders (VAE)~\cite{doersch2016tutorial}.

The training of bimodal generative models that can generate one modality by using the other modality, provided decent results by utilizing variants of VAE referred to as conditional multimodal autoencoders and conditional VAEs~\cite{pandey2017variational,sohn2015learning}. Pu et al.~\cite{NIPS2016_6528} proposed a novel VAE to model images as well as associated labels or captions. They used a Deep Generative Deconvolutional Network (DGDN) as a decoder of the latent image features and deep Convolutional Neural Network (CNN) as an image encoder. The CNN is used to approximate distribution for the latent DGDN features.  

However, these models cannot generate data when they are conditioned interchangeably on both modalities. Vedantam et al.~\cite{DBLP:journals/corr/VedantamFHM17} reported the modification of the VAE to enable a bimodal conditional generation. The method uses a novel training objective and product-of-experts inference network, which can handle partially specified (abstract) concepts in a principled and efficient manner. Perarnau et al.~\cite{perarnau2016invertible} introduced the joint
multimodal VAE (JMVAE), which learns the distribution by using a joint inference network. To
handle missing data at test time, the JMVAE trains $q(z \vert x_1, x_2)$ with two other inference
networks $q(z \vert x_1)$ and $q(z \vert x_2)$.

Kurle et al.~\cite{kurle2018multi} formulated a variational-autoencoder-based framework for multi-source learning, in which each encoder is conditioned on a different information source. This enabled to relate the sources through the shared latent variables by computing divergence measures between individual posterior approximations. 
Wu et al.~\cite{wu2018multimodal} introduced a multimodal VAE (MVAE), which uses a product-of-experts inference network and sub-sampled training paradigm to solve the multimodal inference problem. The model shares parameters to efficiently learn under any combination of missing modalities.

Common for most multimodal methods based on variational inference is the lack of having an expressive family of distributions. This limits the generative power of the models, which we address in our work.

\section{Variational Inference}\label{preliminaries}

The variational inference method approximates intractable probability densities through optimization. Consider a probabilistic model with observations $\mathbf{x}=x_{1:n}$, continuous latent variables $\mathbf{z}=z_{1:m}$, and model parameters $\theta$. In variational inference, the task is to compute the posterior distribution:

\begin{equation}\label{posterior}
    p(\mathbf{z} \vert \mathbf{x},\theta)=\frac{p(\mathbf{z},\mathbf{x} \vert \theta)}{\int_z p(\mathbf{z},\mathbf{x} \vert \theta)}
\end{equation}

The computation requires marginalization over the latent variables $\mathbf{z}$, which often is intractable. In variational methods, a distribution family is chosen over the latent variables with its variational parameters $q(z_{1:m} \vert \phi)$. The parameters that lead to $q$ as close as possible to the posterior of interest are estimated through optimization~\cite{jordan1999introduction}. However, the true posterior often is not in the search space of the distribution family and thus the variational inference provides only an approximation. The similarity between the two distributions is measured by the Kullback–-Leibler (KL) divergence.


Direct and exact minimization of the KL divergence is not possible. Instead, as proposed in \cite{jordan1999introduction}, a lower bound on the log-marginal likelihood is constructed:

\begin{align}
    \log p_\theta(\mathbf{x}) \geq \log p_\theta(\mathbf{x}) &- \mathrm{KL}\left[q_\phi(\mathbf{z}\vert \mathbf{x}), p_\theta(\mathbf{z}\vert \mathbf{x})\right] \label{elbo1} \\ 
    = \E_{q_\phi(\mathbf{z} \vert \mathbf{x})}\left[\lambda \log p_\theta(\mathbf{x}|\mathbf{z})\right]
    &- \beta \mathrm{KL}\left[q_\phi(\mathbf{z}|\mathbf{x}), p(\mathbf{z})\right] \label{elbo2} \\
    &= \mathrm{ELBO(\mathbf{x})} \nonumber
\end{align}

where $\lambda$ and $\beta$ are weights in the Evidence Lower Bound (ELBO). In practice, $\lambda=1$ and $\beta$ is slowly annealed to 1 to form a valid lower bound on the evidence~\cite{bowman2015generating}. In Equation \ref{elbo2}, the first term represents the reconstruction error, while the second corresponds to the regularization. The ELBO function in VAEs is optimized by gradient descent by using the reparametrization trick, while the parameters of the distributions $p$ and $q$ are obtained by neural networks. 





\section{Learning flexible and complex distributions in multimodal variational methods}\label{methods}

Consider a set $\mathbf{X}$ of $N$ modalities, $\mathbf{x}_1, \mathbf{x}_2, \dots, \mathbf{x}_N$. We assume that these modalities are conditionally independent considering the common latent variable $\mathbf{z}$, which means that the model $p(\mathbf{x}_1, \mathbf{x}_2, \dots, \mathbf{x}_N, \mathbf{z})$ can be factorized. Such factorization enables the model to consider missing modalities, as they are not required for the evaluation of the marginal likelihood~\cite{wu2018multimodal}. This can be used to combine the distributions of the $N$ individual modalities into an approximate joint posterior. Therefore, all required $2^N$ multimodal inference networks can be computed efficiently in terms of the $N$ unimodal components. In such a case, the ELBO for multimodal data becomes:

\begin{figure*}[htbp]
\centerline{\includegraphics[scale=0.6]{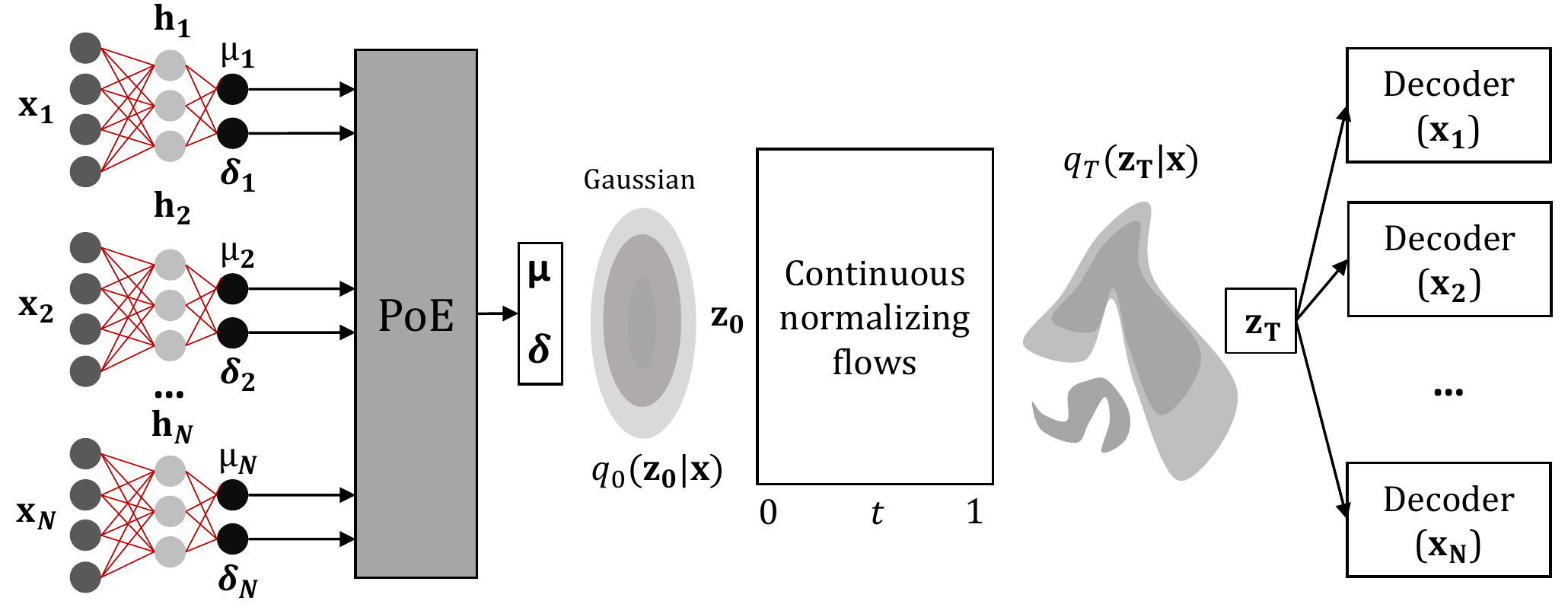}}
\caption{Model architecture of the multimodal VAE with normalizing flows. The model can be used with any type of normalizing flows, transforming $z_0$ to $z_T$.}
\label{modelarchitecture}
\end{figure*}

\begin{equation}
\label{eqelbo}
\begin{aligned}
     \mathrm{ELBO}(\mathbf{X}) = E_{q_\phi(\mathbf{z}\vert\mathbf{X})}\left[\sum_{\mathbf{x}_i \in \mathbf{X}} \lambda_i \log p_\theta(\mathbf{x}_i|\mathbf{z})\right] \\- \beta \mathrm{KL}\left[q_\phi(\mathbf{z}|\mathbf{X}), p(\mathbf{z})\right]
\end{aligned}
\end{equation}

In Figure \ref{modelarchitecture} we present the architecture of the proposed model. Here, the modalities are transformed by encoding inference networks and mapped to a predefined parametric family of distributions, such as the Gaussian. These parameters are combined in the product-of-experts network along with the prior expert~\cite{hinton2002training}. The output of the product-of-experts is again a parametric distribution, which has low expressive power and is not sufficient for good approximations to the true posterior.

\begin{figure}[htbp]
\centerline{\includegraphics[scale=1]{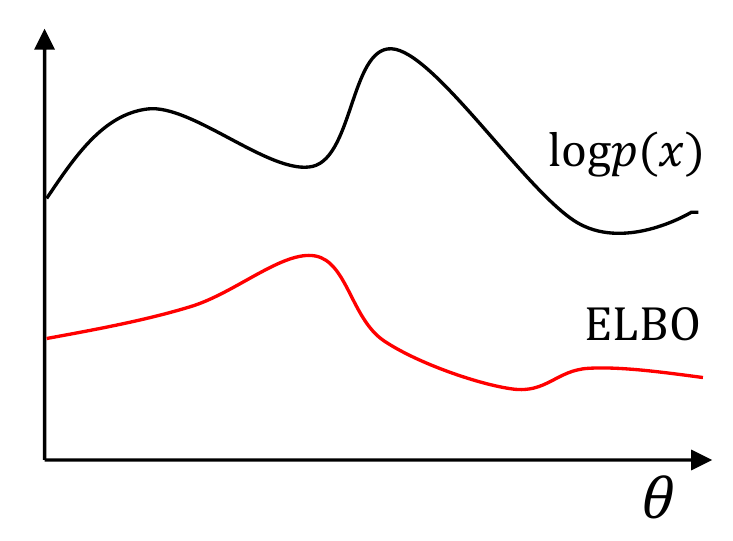}}
\caption{Difference between the ELBO and true log likelihood. With the increase in their difference, their local maxima tend to differ
more, which leads to a bias in the optimization of the model parameters.}
\label{fig:elbo}
\end{figure}

A better variational approximation to the posterior would provide better values for the ELBO. As the ELBO is only a lower bound on the marginal log-likelihood, they do not share the same local maxima. Therefore, if the lower bound is too far off from the true log-likelihood, the maxima of the functions can differ even more, as illustrated in Figure~\ref{fig:elbo}. Due to its simplicity, one of the most commonly used variational distributions for the approximate posterior $q_\phi(\mathbf{z} \vert \mathbf{x})$ is the diagonal-covariance Gaussian. However, with such a simple variational distribution, the approximate posterior will most likely not match the complexity of the true posterior. According to Equation \ref{elbo1}, this can lead to a larger KL divergence term and thus a larger difference between the ELBO and true log-likelihood, which often leads to a low-performance generative model.

For the optimal variational distribution that provides $KL=0$, $q$ matches the true posterior distribution. This can be achieved only if the true posterior is a part of the distribution family of the approximate posterior; therefore, the ideal family of variational distributions $q_\phi(\mathbf{z} \vert \mathbf{x})$ is a highly flexible one.

An approach to achieve such flexible and complex distributions is to utilize normalizing flows~\cite{tabak2013family}. Consider $\mathbf{z} \in \mathbb{R}$ as a random variable and $f: \mathbb{R}^d \rightarrow \mathbb{R}^d$ as an invertible mapping. The function $f$ can transform $\mathbf{z} \sim q_\phi(\mathbf{z})$ and yield a random variable $y=f(\mathbf{z})$, which has the following probability distribution:

\begin{equation}\label{cov}
    q_y(\mathbf{y}) = q_\phi(\mathbf{z})\left|\det\frac{\partial f^{-1}}{\partial \mathbf{z}}\right|
\end{equation}

This transformation is also known as the change-of-variable formula. Composing a sequence of such invertible transformations to a variable is known as normalizing flows:

\begin{equation}\label{eq:seq_transform}
    \mathbf{z}_{t+1}=\mathbf{z}_t+f(\mathbf{z}_t,\theta_t)
\end{equation}
where $t\in \{0, \dots, T\}$ and $\mathbf{z}_t \in \mathbb{R}^D$.

Generally, the main bottleneck of the use of the change-of-variables formula is the computation of the determinant of the Jacobian, which has a cubic cost in either the dimension of $\mathbf{z}$ or number of hidden units.

Rezende et al. introduced a normalizing flow, referred to as planar flow, for which the Jacobian determinant could be computed efficiently in $O(D)$ time~\cite{rezende2015variational}. Berg et al.~\cite{berg2018sylvester} introduced Sylvester normalizing flows, which can be regarded as a generalization of planar flows. Sylvester normalizing flows remove the well-known single-unit bottleneck from planar flows, making the single transformation considerably more flexible.

It is simple to sample from these models and they can be trained by maximum likelihood by using the change-of-variables formula. However, this requires placing awkward restrictions on their architectures, such as partitioning of dimensions or use of rank-one weight matrices, to avoid a $O(D^3)$ cost determinant computation. These limitations are overcome with the introduction of continuous normalizing flows\cite{grathwohl2019ffjord}, which we refer to in the following.

The sequence of transformations shown in Eq.~\ref{eq:seq_transform} can be regarded as an Euler discretization of a continuous transformation~\cite{chen2018neural}. If more steps are added, in the limit, we can parametrize the transformations of the latent state by using an ordinary differential equation (ODE) specified by a neural network:
\begin{equation}
    \frac{\partial \mathbf{z}(t)}{\partial t} = f(\mathbf{z}(t), t, \theta)
\end{equation}
where $t$ is now continuous, i.e. $t\in \mathbb{R}$. Chen et al.\cite{chen2018neural} reported that the transfer from a discrete set of layers to a continuous transformation simplifies the computation of the change in normalizing constant. In the generative process, sampling from a base distribution $\mathbf{z_0} \sim p_{\mathbf{z_0}}(\mathbf{z_0})$ is carried out. Considering the ODE defined by the parametric function $f(\mathbf{z}(t), t; \theta)$, the initial value problem:
\begin{equation}
    \mathbf{z}(t_0) = \mathbf{z_0}, \frac{\partial \mathbf{z}(t)}{\partial t} =
f(\mathbf{z}(t), t; \theta)
\end{equation}
is solved to obtain $\mathbf{z}(T)=\mathbf{z}_T$, which constitutes the transformed latent state. The change in log-density under the continuous normalizing flow (CNF) model follows a second differential equation, referred to as formula of instantaneous change of variables:
\begin{equation}
    \frac{\partial \log q_t(\mathbf{z}(t))}{\partial t} = -\mathrm{Tr}\left(\frac{\partial f}{\partial \mathbf{z} (t)}\right)
\end{equation}
We can compute the total change in log-density by integrating over time:
\begin{equation}
    \log q_T(\mathbf{z}(T)) = \log q_0(\mathbf{z}(t_0)) - \int_{t_0}^{T} \mathrm{Tr}\left(\frac{\partial f}{\partial \mathbf{z}(t)} 
    \mathrm{d} t \right)
\end{equation}
Extending Chen et al.~\cite{chen2018neural}, Grathwohl et al. introduced the Hutchinson's trace unbiased stochastic estimator of the likelihood, which has $O(D)$ time cost, which enables completely unrestricted architectures~\cite{grathwohl2019ffjord} .

Using such continuous normalizing flows, we transform the joint distribution $\mathbf{z}_0$ to $\mathbf{z}_K$ to obtain a more complex distribution, as shown in Figure \ref{modelarchitecture}. Lastly, $\mathbf{z}_K$ is used to reconstruct the original inputs $\mathbf{x}_1, \dots, \mathbf{x}_N$ through the decoder networks.

Utilizing this, we derive new KL divergence in the multimodal setting. With this modification, the new ELBO (Equation \ref{eqelbo}) has new KL divergence, which is computed by:

\begin{equation}
\label{klnew}
\begin{aligned}
    \mathrm{KL} \sim \log q_T(\mathbf{z}_T \vert \mathbf{X}) - \log p(\mathbf{z}_T) = \\ 
    \log q_0(\mathbf{z}_0 \vert \mathbf{X}) - \log p(\mathbf{z}_T) - \int_{0}^{T} \mathrm{Tr} \left( \frac{\partial f}{\partial{\mathbf{z}_t}} \mathrm{d}t \right)
    \end{aligned}
\end{equation}

We obtain the new objective of the multimodal VAE by replacing the KL divergence term in Equation \ref{eqelbo} with that in Equation \ref{klnew}. Training the model with the new objective gives larger generative power to model, which can be utilized to improve the learning of joint distributions in complex multimodal data.

\section{Evaluation}\label{evaluation}
In this section, we present multiple experiments carried out to evaluate our model performance which we name multimodal variational autoencoder with continuous normalizing flows (MVAE-CNF) on several datasets and learning tasks including MNIST, FashionMNIST, KMNIST, EMNIST, CelebA, and Flickr. 

The MNIST-like unimodal datasets are transformed into multimodal datasets by treating the labels as a second modality, which has been also carried out in previous related work. We refer to the images as the image modality and labels as the text modality. For each of them, the comparison is carried out against the state-of-the-art MVAE. In \cite{wu2018multimodal}, MVAE is compared to previous multimodal variational approaches.

For a proper evaluation, the same model architectures for the encoders and decoders, learning rates, batch sizes, number of latent variables, and other parameters are used. 

Our model, MVAE-CNF, has additional parameters for the CNFs. We used the version from~\cite{grathwohl2019ffjord} as the CNF without amortization. As the solver for the ODE, we used the Euler method to achieve computational efficiency and demonstrate that even with the utilization of the simplest solver, the method is improved compared to the current state of the art. For the CNF settings, we used only one block of CNFs, 256 squash units for the layers of ODENet, and softplus non-linearity as the activation function for the ODENet.

For the four MNIST datasets, we used linear layers having 512 units for the encoders and decoders for both modalities. Each of the encoders is followed by $\mu\text{ and }\sigma$ hidden layers having 128 units with Swish non-linearities. To optimize the lower bound in these data, we used annealing where the factor of the KL divergence is linearly increased from 0 to 1 during 20 out of the total of 60 epochs\cite{bowman2015generating}. 

For the image transformation tasks, we used the CelebA dataset. For the encoders and decoders, we used simple convolutional networks composed of 4 [convolution-batch normalization] layers, whose last layer is flattened and decomposed into $\mu$ and $\sigma$ with 196 units for all models. We slowly anneal the factor of the KL from 0 to 1 during 5 epochs and train for 20 epochs owing to the limitations of the available computational resources. The above CNF parameters are unchanged.

Lastly, the FLICKR dataset is composed of images and textual descriptions. We used the image and text descriptors as in \cite{srivastava2012multimodal}. All parameters of the decoders and encoders, batch size, epochs, learning rates, etc. are similar to those of the MNIST-like datasets.


\begin{figure}[!htb]
\minipage{0.47\textwidth}
  \includegraphics[width=\linewidth]{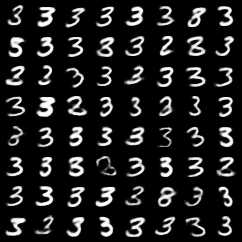}
  \caption{Generated samples from MNIST, sampled from the conditional distribution, when $\mathbf{X}_2=3$.}\label{mnistgeneration}
\endminipage\hfill
\minipage{0.47\textwidth}
  \includegraphics[width=\linewidth]{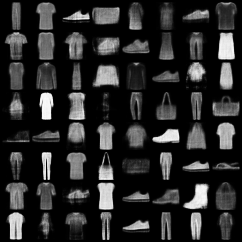}
  \caption{Generated samples from the Fashion MNIST dataset, sampled from the unconditioned joint distribution.}\label{fmnistgeneration}
\endminipage
\end{figure}

\subsection{Evaluation of the model and generated sample quality}

For the MNIST-like datasets, we denote the images as $\mathbf{X}_1$ and labels as $\mathbf{X}_2$ and we set $\lambda_1$ = 1 and $\lambda_2$ = 50 (Equation \ref{eqelbo}). The up-weighting of the reconstruction error for the low-dimensional modalities is important for the learning of good joint distribution. 

Table \ref{mnistexp} shows the joint ELBO loss, computed according to \ref{eqelbo} and \ref{klnew}, ELBO losses for both modalities separately, and binary cross-entropy (BCE) values for $\mathbf{X}_1$ and $\mathbf{X}_2$. Lower values for all columns imply better performance. We present the results for the four MNIST datasets. For both modalities, jointly and separately, the MVAE-CNF outperforms the MVAE in ELBO loss. As presented in the table, we evaluated the BCEs for both modalities, by ignoring the KL divergence, which provides a direct comparison of the models in terms of learning of the reconstructions. The MVAE-CNF outperforms the MVAE in all datasets. This is due to the increased capacity of the joint distribution because of the utilization of continuous normalizing flows.

\begin{table*}
\caption{Binary cross-entropy and ELBO losses for four MNIST-like datasets}
\centering
\resizebox{.99\linewidth}{!}{%
\begin{tabular}{l|l|l|l|l|l|l|l|l|l|l}
\hline
\multirow{2}{*}{} & \multicolumn{2}{c|}{$\mathbf{X}_1$ BCE loss} & \multicolumn{2}{c|}{$\mathbf{X}_2$ BCE loss} & \multicolumn{2}{c|}{ELBO joint} & \multicolumn{2}{c|}{ELBO $\mathbf{X}_1$} & \multicolumn{2}{c}{ELBO $\mathbf{X}_2$} \\ \cline{2-11} 
                  & MVAE          & MVAE-CNF       & MVAE         & MVAE-CNF        & MVAE         & MVAE-CNF         & MVAE         & MVAE-CNF        & MVAE         & MVAE-CNF        \\ \hline
MNIST             & 80.9758       & 78.2835        & 0.0142       & 0.0078          & 100.9352     & 98.0086          & 99.5205      & 96.9436         & 5.0624       & 3.8717          \\
FashionMNIST      & 225.6746      & 224.4749       & 0.034        & 0.0102          & 240.9705     & 239.4592         & 238.1776     & 237.4313        & 4.1957       & 3.9981          \\
KMNIST            & 188.2816      & 182.3637       & 0.0072       & 0.070           & 214.2033     & 209.6676         & 210.6992     & 206.7195        & 3.8445       & 3.7986          \\
EMNIST(letters)   & 112.2935      & 111.8147       & 0.0064       & 0.0062         & 143.9629     & 141.6544       & 138.7585     & 137.6635          & 5.4532         & 5.5672              \\ \hline
\end{tabular}
}

\label{mnistexp}
\end{table*}


Furthermore, we evaluated the qualities of the generated samples for MNIST, FashionMNIST, and KMNIST. First, we trained a simple neural network for classification. The classifier was not tuned for the best accuracy; it served only for the relative comparison of the generative models. Subsequently, from each of the MVAE and MVAE-CNF, we generated 1000 image samples conditioned on the labeled modality and utilized the classifier to predict the appropriate class. We summarize the accuracies of the predictions by using the generated data in Table \ref{qualitygenerated}. Notably, the MVAE-CNF can generate samples with approximately 10–50\% higher qualities.

In Figures \ref{mnistgeneration} and \ref{fmnistgeneration}, we illustrate the generated samples from the MNIST and FashionMNIST datasets. We find the samples to be good quality, and find conditional samples to be largely correctly
matched to the target label. Figure \ref{mnistgeneration} shows images sampled from the conditional distribution, while Figure \ref{fmnistgeneration} shows generated samples from the joint distribution of the FashionMNIST dataset.

\begin{table}
\parbox{.46\linewidth}{
\caption{Quality of the generated samples evaluated by a supervised classifier}
\centering
\begin{tabular}{l|r|r}
\hline
Dataset      & MVAE & MVAE-CNF \\ \hline
MNIST        & 0.67     & 0.73         \\ 
FashionMNIST & 0.24    & 0.44        \\ 
KMNIST       & 0.11     & 0.26         \\ \hline
\end{tabular}
\label{qualitygenerated}
\caption{Learning with few examples}
\centering
\begin{tabular}{l|r|r|r}
\hline
Dataset      & Image loss & Text loss & Joint loss \\ \hline
Flickr 15\%  & 1241.7388           & 23.5962           & 2571.3684            \\ 
Flickr 30\%  & 1229.2150           & 23.8598           & 2544.5045            \\ 
Flickr 100\% & 1214.1170           & 23.2236           & 2535.6338            \\ \hline
\end{tabular}
\label{flickr}
}
\hfill
\parbox{.46\linewidth}{
\caption{Image transformations on celebA}
\centering
\begin{tabular}{l|l|r}
\hline
Cross-entropy & MVAE   & MVAE-CNF \\ \hline
IG: Image      & 6389.5317  &6322.8523          \\ 
IG: Gray      & 2239.9395        &2197.7927          \\ 
IG: Image-Gray       & 8548.3848     &8478.5559        \\ \hline
IE: Image       & 6446.3184        &6315.6718           \\ 
IE: Edge       & 1005.5892       & 992.2274         \\ 
IE: Image-Edge   & 7423.9004       & 7289.6343          \\ \hline
IM: Image  & 6396.6943        & 6329.7452         \\ 
IM: Mask  & 245.4223       & 239.4394          \\ 
IM:  Image-Mask  & 6632.7891       & 6556.8282          \\ \hline
\end{tabular}
\label{imtransforms}
}
\end{table}

\subsection{Image Transformation Tasks}

In this section, we compare the performances of the multimodal generative models on an image transformation task. In edge detection, one modality is the original image, while the other is the image with the edges. In colorization, the two modalities are the same image in color and grayscale. Similarly to the approach in \cite{wu2018multimodal}, we apply transformations to images from the CelebA dataset. For colorization, we transform the RGB images to grayscale, for edge detection we use the Canny detector, and for landscape masks, we use dlib and OpenCV.

We used the validation subset of CelebA as training data and test subset for the computation of the scores. Both MVAE and MVAE-CNF use the same parameters, as explained above, with $\lambda_i=1$.

\begin{figure}
\centerline{\includegraphics[height=0.45\textwidth]{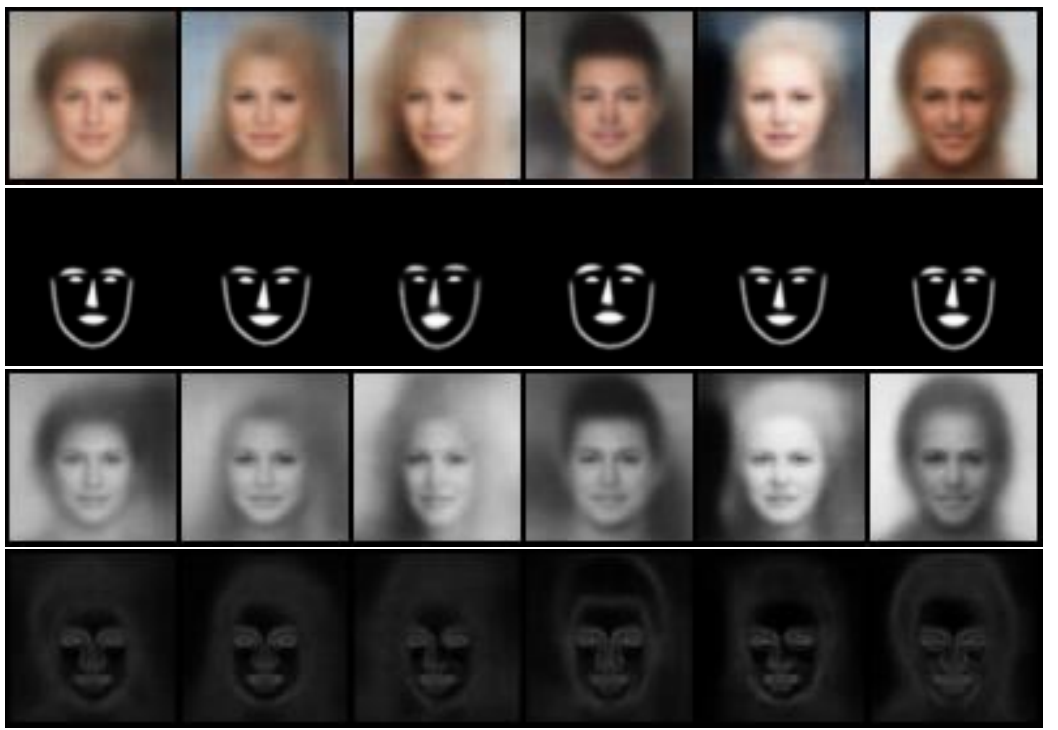}}
\caption{Image transformations generated by using the joint distribution. From top to bottom: colorization, masks, grayscale, and edge extraction.}
\label{transforms}
\end{figure}

In Table \ref{imtransforms}, we show the results and comparison of the different image transformation tasks. The MVAE-CNF model outperforms MVAE in all tasks. We show the results for the cross-entropy loss for the joint image-gray, image-edge, and image-mask tasks and each of the modalities separately. Furthermore, in Figure \ref{transforms}, we show the unconditional sampling from the joint distribution. The samples preserve the main task. However, they are characterized as samples from other variational models with smooth and blur effects. On CelebA, the result suggest that the product-of-experts approach generalizes to a
larger number of modalities (19 in the case of CelebA), and that jointly training shares statistical strength.

In Table \ref{flickr}, on the Flickr dataset, we show that even with 5\% of matched data samples, the obtained results almost reach the same performance like that when all data are supervised. This implies that the model can learn the data with only a few supervised samples and utilize the unmatched unlabeled modalities.

\section{Conclusion and Future Work}\label{conclusion}

Modeling the joint distribution of multimodal data requires larger expressive power to capture high-level concepts and provide better data representations. Previous multimodal generative models based on variational  inference  are  limited  due  to  the  lack  of  flexibility  of  the  approximate posterior,  which  is  obtained  by  searching  within  a  known  parametric  family of distributions. We presented a new approach, which aims to improve the family of multimodal variational models. We utilized the product-of-experts as the inference network to approximate the joint data distribution, which is followed by continuous normalizing flows, which transform the parametric distribution into a more complex distribution. This provides a larger expressive power of the model. In general, these flows can be used for multimodal generative models of any architecture.

Through several experiments, we showed that our approach outperforms the state of the art on variety of tasks including the learning of joint data distributions, image transformations, weakly supervised learning, and generation of missing data modalities. Utilizing a trained classifier on the generated samples, we also show that learning more powerful approximate joint distributions improves the quality  of the generated  samples by more than 0.15\% compared to the previous multimodal VAE methods. As future work, in support of our results, we believe that the focus should lie on efficiently obtaining better and more complex joint distributions that can learn the underlying data.

\section*{Acknowledgments} \label{sec:ACKNOWLEDGMENTS}
This work was funded by the German Ministry for Education and Research as BIFOLD - Berlin Institute for the Foundations of Learning and Data (ref. 01IS18025A and ref 01IS18037A). We also acknowledge financial support by BASF under Project ID 10044628.

\bibliographystyle{splncs04}
\bibliography{sample}
\end{document}